\documentclass[twoside,11pt]{article}

\usepackage{jmlr2e}
\usepackage{hyperref}       
\usepackage{cleveref}
\usepackage{url}            
\usepackage{booktabs}       
\usepackage{amsfonts}       
\usepackage{nicefrac}       
\usepackage{microtype}      
\usepackage{graphicx}
\usepackage{amsmath,cases,amssymb}
\usepackage{mathrsfs}
\usepackage{algorithm}
\usepackage{caption}
\usepackage{algorithmic}
\usepackage{bm}
\usepackage{multirow}
\usepackage{mathrsfs}
\usepackage{wrapfig}
\usepackage{url}
\usepackage{breakurl}
\usepackage{amsmath}
\usepackage{amssymb}

\firstpageno{1}

\begin{document}

\title{Dual Pattern Learning Networks by Empirical Dual Prediction Risk Minimization}

\author{\name Haimin Zhang \email Haimin.Zhang@student.uts.edu.au \\
        \name Min Xu  \email Min.Xu@uts.edu.au \\
       \addr Faulty of Engineering and IT\\
       University of Technology Sydney\\
       81 Broadway, Ultimo, NSW 2007\\
        Australia
       }

\editor{}

\maketitle

\begin{abstract}
  Motivated by the observation that humans can learn patterns from two given images at one time, we propose a dual pattern learning network architecture in this paper.
  Unlike conventional  networks, the proposed architecture has two input branches and two loss functions.
  Instead of minimizing the empirical risk of a given dataset, dual pattern learning networks is trained by minimizing the empirical dual prediction loss.
  We show that this can improve the performance for single image classification.
  This architecture forces the network to learn  discriminative class-specific features by analyzing and comparing two input images.
  In addition, the dual input structure allows the network to have a considerably large number of image pairs, which can help address the overfitting issue due to limited training data.
  Moreover, we propose to associate each input branch with a random interest value for learning corresponding image during training.
  This method can be seen as a stochastic regularization technique, and can further lead to generalization performance improvement.
  State-of-the-art deep networks can be adapted to dual pattern learning networks without increasing the same number of parameters.
  Extensive experiments on  CIFAR-10, CIFAR-100, FI-8, Google commands dataset, and MNIST demonstrate that our DPLNets exhibit better performance than original networks.
  The experimental results on subsets of CIFAR-10, CIFAR-100, and MNIST demonstrate that dual pattern learning networks have good generalization performance on small datasets.
  
\end{abstract}


\section{Introduction}

\normalsize
Deep convolutional neural networks (CNNs) have proven to be powerful machine learning models.
CNNs have achieved great success in visual recognition tasks such as image recognition (\cite{krizhevsky2012imagenet, he2016deep, sabour2017dynamic}), object detection (\cite{lin2017feature, du2017spatio, he2017mask}), and video classification (\cite{shen2017fast, wang2017spatiotemporal}).
Integrating feature learning and classifiers in an end-to-end manner, deep networks can learn features automatically from data without human involvement during training. It has been shown that features learned by CNNs are much discriminative compared with hand-crafted features (\cite{krizhevsky2012imagenet}), and that  features extracted from a CNN pretrained on a large scale dataset  can be transferred to other visual recognition tasks (\cite{donahue2014decaf}).

Researchers have spent much effort in designing CNN architectures to improve the performance of CNNs.
For example, He \emph{et al.}  proposed the residual learning framework (\cite{he2016deep}).
This framework eases the training of deep networks, and enables them to be considerably deep.
Residual networks (ResNets) have led to performance improvement in both visual and non-visual tasks.
Huang \emph{et al.}  proposed densely connected networks (DenseNets) (\cite{huang2017densely}). In DesnseNets,  each layer is connected to every other layer in a feed-forward fashion.
This architecture substantially reduces the number of parameters, and is highly computationally efficient as a result of feature reuse.
State-of-the-art deep networks usually consist of many layers with a large number of parameters.
For example, the ResNeXt-19 ($8\times64$d) (\cite{xie2017aggregated}) contains approximately $3 \times 10^7$ parameters to model the $5\times 10 ^4$ images in CIFAR-10 (\cite{krizhevsky2009learning}).  The VGGNet-16 (\cite{simonyan2014very}) has around $10^8$ parameters to model the $10^6$ images in ImageNet (\cite{russakovsky2015imagenet}).
The larger number of parameters makes deep models prone to overfitting; therefore, training deep networks requires huge amounts of data.
However, collecting data and labeling them are laborious work, especially when domain experts are necessary to distinguish between fine-grained visual categories. It is extremely difficult to collect training samples for some tasks.

\begin{figure}[!t]

  \begin{center}
    \includegraphics[width=0.618\textwidth]{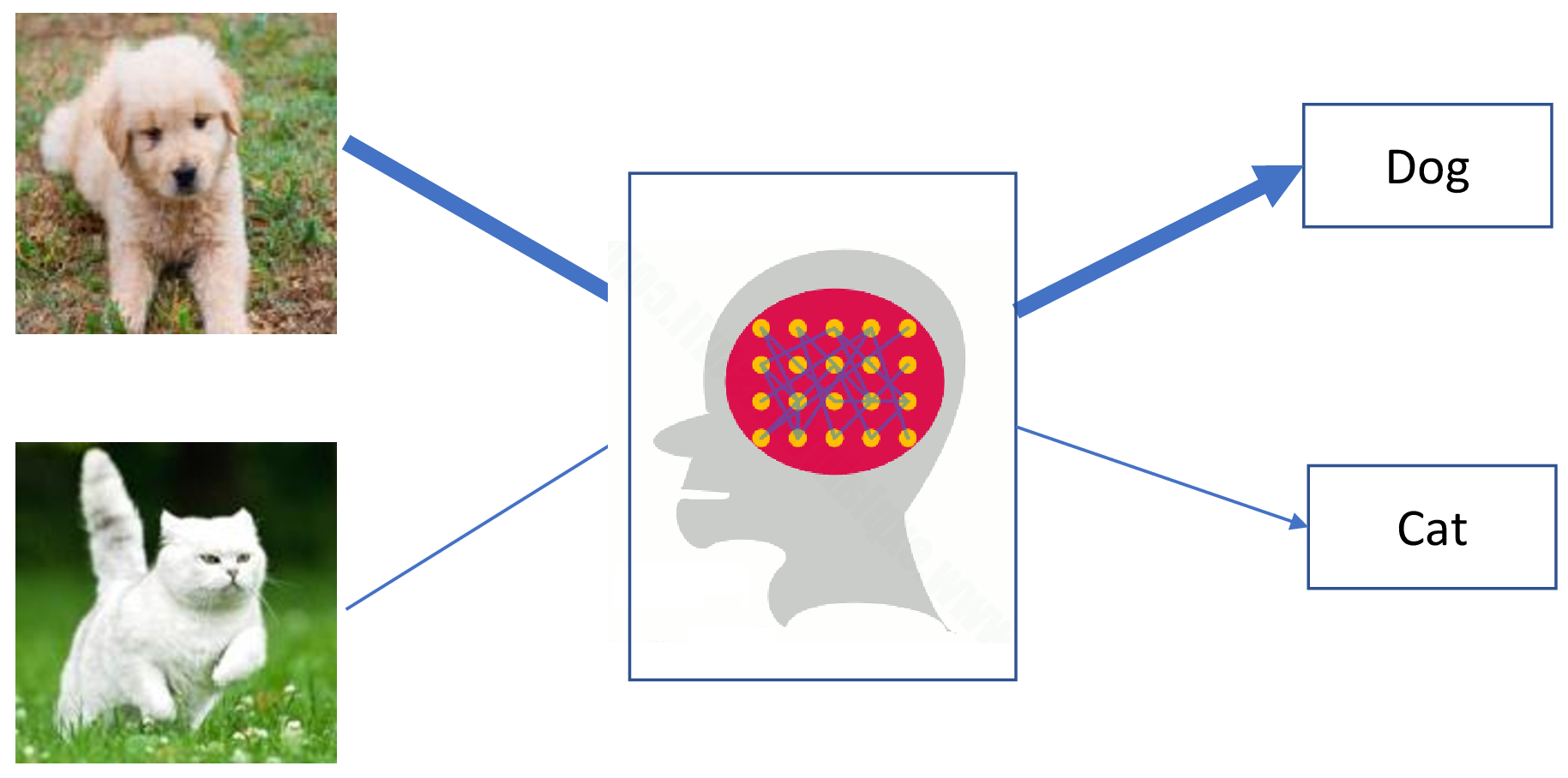}
  \end{center}
  \caption{Humans can learn patterns from two given images at one time. They may have more interest in learning one image than the other image when learning from two images. In this figure, the human is more interested in learning dog at this time (boldness of lines indicates interest value).}
  \label{fig:dualPatternLearning}
\end{figure}

One commonality most existing deep networks share is that they have a single input branch, and are trained by minimizing the empirical error of a given training dataset using an optimization method such as stochastic gradient descent (SGD) and Adam (\cite{kingma2014adam}).
Unlike neural networks, humans have a strong ability to perceive and recognize new patterns.
In particular, humans can  learn knowledge from two given images at one time. An illustration is shown in Figure \ref{fig:dualPatternLearning}.
During the learning process, people learn class-specific features that can be used for both recognizing and differentiating the two given images.
Inspired by this observation, we propose a dual pattern learning (DPL) network architecture in this paper.

In the proposed DPLNet  architecture, we design two input branches and two loss functions in order to do dual pattern learning.
Two input images are processed by the two  branches in parallel, and feature maps generated by the two branches are then fused together to backbone network.
This architecture  forces the network to learn robust image features that can be used to both recognize and differentiate the two input images.
The two input branches share the same parameters, which ensures that features learned by the two branches are consistent.
The DPLNet is trained by minimizing the average of empirical prediction errors for two inputs.
We show that  dual pattern learning networks are powerful for image recognition.
Unlike the Siamese architecture (\cite{koch2015siamese}), which consists of two identical networks with the same weights, the proposed DPLNet is trained to classify input images instead of learning a similarity between two input images.

People may have more interest in  one image than the other image when given two images to learn.
This might be due to reasons such as personal preference or prior knowledge.
Inspired by this observation, we propose to associate each input branch with a random interest value for learning corresponding image during training.
The DPLNet pays more attention to the image which it has more interest in.
This method can further improve the generalization performance of the proposed network.

%


This paper provides the following three contributions:
\begin{itemize}
  \item We propose a dual pattern learning network architecture which is trained by minimizing the  empirical dual prediction loss of a training dataset.  This architecture enables the network to learn discriminative class-specific features by analyzing and comparing two input images.  The dual input structure enables the network to have a  large number image pairs to train the network. This can help deal with the overfitting issue due to lack of training data.
  \item We propose to associate each input branch with a random interest value for learning corresponding image during training.  This method can be seen as a stochastic regularization technique, and can further lead to generalization performance improvement.
This value can be considered as an interest value for learning the corresponding image.
This technique can improve the generalization performance.
  \item State-of-the-art deep networks can be easily adapted to dual pattern learning networks (DPLNets) without increasing the number of  parameters.We evaluate DPLNets on five benchmark dataset, \emph{i.e.,}  CIFAR-10, CIFAR-100,  FI-8, Google commands dataset, and MNIST, wherein they lead to performance improvement compared with original networks. The experimental results on subsets of CIFAR-10, CIFAR-10, and MNIST demonstrate that our DPLNets have good generalization performance on small datasets.
\end{itemize}

The rest of the paper is organized as follows. Section \ref{sec:related_work} reviews related work on deep neural networks. The proposed DPLNet architecture is introduced in section \ref{sec:method}. Experimental results are presented and
discussed in section \ref{sec:experiments}. Finally, we conclude this paper in section \ref{sec:conclusion}.

\section{Related Work} \label{sec:related_work}
CNN was originally introduced  by Lecun \emph{et al.} (\cite{lecun1989backpropagation}) in late 1980s. Development in computer hardware and network structures made training deep neural networks on large scale dataset, such as ImageNet  (\cite{russakovsky2015imagenet}), feasible only recently.  
In 2012, Krizhevsky \emph{et al.} (\cite{krizhevsky2012imagenet}) proposed AlexNet.
AlexNet consists of five convolutional layers and three fully connected layers.
This is the first CNN proposed for large scale image  classification.
AlexNet achieved superior performance compared with hand-crafted features.

Since the introduction of AlexNet, many CNN architectures have been developed to improve performance.
These work include exploring increasing the depth (the number of layer) and the width (the number of channels in each layer).
For example, Simonyan \emph{et al.}  (\cite{simonyan2014very}) investigated the effect of the  network depth on its accuracy, and proposed VGGNets with 16 and 19 layers.
He \emph{et al.} introduced identity shortcut connections, and proposed ResNets in (\cite{he2016deep}).
This architecture makes very deep networks easy to optimize.
ResNets have achieved performance improvement for many tasks.
Huang \emph{et al.} explored the width of networks, which refers to the number of channels in a layer, and proposed DenseNets in (\cite{huang2017densely}).
For each layer in DenseNets, the feature-maps of all preceding layers are
used as inputs. The DenseNet architecture encourages feature reuse, and considerably reduces the number of parameters.
In addition to depth and width, cardinality, which refers to the size of the set of transformations, has also been researched.
Xie \emph{et al.} proposed ResNeXt in (\cite{xie2017aggregated}).
This network is constructed by repeating a building block that aggregates a set of transformations with the same topology.
They showed that increasing cardinality is more effective than going deeper or increasing the width.

Training deep neural networks requires huge amounts of data to reduce overfitting because they have a large number of parameters.
Methods include stopping training as soon as performance on a validation set starts to get worse, introducing weight penalties of various
kinds such as L1 and L2 regularization and soft weight sharing (\cite{nowlan1992simplifying}).
To reduce over-fitting on training data, the easiest  method is to enlarge the dataset using label-preserving transformations (\cite{krizhevsky2012imagenet}).
Commonly used data augmentation methods include random crop, color jittering, horizontal/vertical flip of images.
Recently, researcher started to use generative adversarial networks  to generate samples for data augmentation. For example, Zheng  \emph{et al.} proposed to use adversarial samples to improve person re-identification baselines (\cite{zheng2017unlabeled}).
The authors of (\cite{Xie_2017_ICCV}) proposed to use adversarial samples  semantic segmentation and object detection.

In addition, many methods have  been developed to improve the generalization performance of deep networks.
For example, Srivastava \emph{et al.}  proposed a method referred to as  dropout (\cite{srivastava2014dropout}). The key idea of dropout is to randomly drop units of the neural network during training.
The neurons which are dropped out in this way do not contribute to the forward pass and do not participate in back-propagation.
This technique forces networks to learn more robust features that are useful in conjunction with many different random subsets of the other neurons.
Ioffe \emph{et al.}  proposed the batch normalization method (\cite{ioffe2015batch}). This method draws its strength from making normalization a part of the model architecture and performing the normalization for each training mini-batch.  Batch normalization allows us to use much higher learning rates and be less careful about parameters initialization.

\section{The Proposed DPLNet Achitecture} \label{sec:method}

\subsection{Empirical Risk Minimization Revisit}
Let $X\in {\rm I\!R}^p$ denote a random input vector, and $Y \in {\rm I\!R} $ a random output variable.
In conventional supervised learning theory, we aim to find a function $f(X) \in \mathcal{F}$ for predicting $Y$ given values of the input $X$.
It is assumed that $(X,Y)$ follows a joint distribution $P(X,Y)$.
This theory requires a loss function $\ell(f(X), Y)$ for penalizing errors between prediction $f(\bm{x})$ and actual target $y$.
We minimize the expected prediction error (EPE), which is also known as expected risk, for choosing $f$.
The EPE is given as follows:

\begin{equation} \label{eq:supervised}
    \begin{aligned}
    {\rm EPE}(f) = \int\ell(f(x), y){\rm d}P(X,Y).
    \end{aligned}
\end{equation}

Unfortunately, the distribution $P$ is implicit for most applications. Instead, we have access to a training dataset  $\mathcal{D}=\{({x}_i,{y}_i)\}_{i=1}^{N}$ with $N$ labelled training data, where ${x}_i$ and $y_i$ represent the $i$-th data and its corresponding label, respectively.
Using the dataset $\mathcal{D}$, the distribution $P$ could be approximated by the empirical distribution:

\begin{equation} \label{eq:emperical_dist}
    \begin{aligned}
    P_{\delta}(x, y) =  \frac{1}{N}\sum_{i=1}^{N}\delta(x = x_i, y = y_i),
    \end{aligned}
\end{equation}
where $\delta(x=x_i,y=y_i)$ indicates a Dirac mass centered at $(x_i, y_i)$.
The Dirac delta function $\delta(x=x_i,y=y_i)$ is  defined such that it is zero-valued everywhere except
$(x_i, y_i)$, yet integrates to 1. With the empirical distribution $P_{\delta}$, we can approximate the expected risk by the empirical risk:

\begin{equation} \label{eq:erm}
    \begin{aligned}
    R(f) = \frac{1}{N}\sum_{i=1}^{N}\ell(f(x_i), y_i).
    \end{aligned}
\end{equation}

Learning the prediction function $f$ by minimizing Equation (\ref{eq:erm}) is know as the empirical risk minimization (EPM) (\cite{vapnik1998statistical}) principle.

\subsection{Dual Pattern Learning}

The proposed dual pattern learning framework intends to learn from dual inputs.
Unlike $f$ for empirical risk minimization, the prediction function $g$ for dual pattern learning takes two samples as input, and is obtained by minimizing the following risk:

\begin{equation} \label{eq:supervised}
    \begin{aligned}
    R_{DPL}(f) = \frac{1}{N}\sum_{i=1}^{N}\ell(g(x_i, x_j), y_i, y_j), \;\; 1 \le j \le N, 
    \end{aligned}
\end{equation}
where the loss function is to be defined to penalize errors for dual prediction results.
The risk $R_{DPL}$ is referred to empirical dual prediction risk in this paper.

Deep neural networks consist of many nonlinear hidden layers. The extreme nonlinearity  makes them powerful to learn complicated relationships between their inputs and outputs.
The prediction function $g$ is represented by a deep neural network in this work.
We show that dual pattern learning networks trained by minimizing the average of dual prediction errors of a given dataset improves the performance for single image classification.

\begin{figure*}[!t]
\begin{center}
\includegraphics[width=\linewidth]{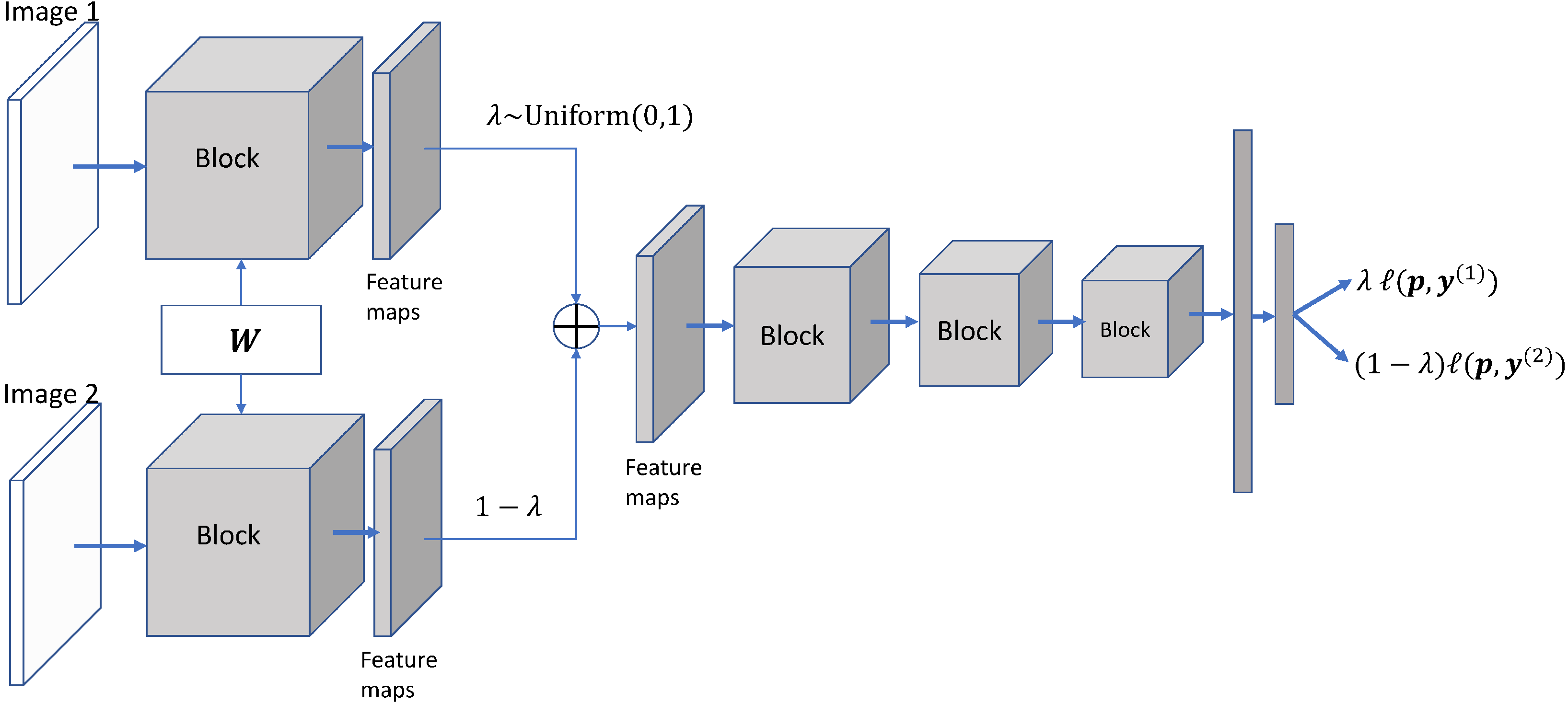}
\end{center}
   \caption{An illustration of the proposed DPLNet architecture. This architecture has two input breaches and two loss functions. The two input branches share the same parameters. Two input images are processed by the two branches in parallel. Feature maps generated by the two branches are fused together to backbone network. We perform weighted fusion. A value $\lambda$ is sampled from the standard uniform distribution as weight for one branch, and $1-\lambda$ for the other branch. The weight associated with each branch can be considered as an interest value for learning the corresponding image. The two interest values are also assigned to the two loss functions.}
\label{Fig:overview}
\end{figure*}

An illustration of the proposed DPLNet architecture is shown in Figure \ref{Fig:overview}. It contains several blocks, each consists of a number of convolutional layers. Feature maps generated within the same block have the same height and width.
We design two input branches and two loss functions  to simulate the human dual pattern learning process.
Two input images are processed by the two  branches in parallel, and feature maps generated by the two input branches are then fused together to backbone network which ends with a fully connected layer with softmax.
The two input branches share the same parameters.
This guarantees that the two input branches generate consistent feature maps for two input images, which means that feature maps generated by the two  branches are the same regardless of input orders.
This architecture forces the network to learn discriminative class-specific features by analyzing and comparing two input images; therefore,  DPLNets encourage learned features  to  have large inter-class margins compared with conventional networks.
We use random weighted combination of feature maps generated by the two input branches as input to backbone network during training.
In particular, we randomly sample a value $\lambda$ from the standard uniform distribution, as given below:
\begin{equation} \label{eq:beta}
    \begin{aligned}
     \lambda \sim Uniform(0, 1).
    \end{aligned}
\end{equation}

The value of $\lambda$ is used as weight for one branch, and $1-\lambda$ for the other branch.
The two weights can be considered as interest values for learning corresponding images.
The fused feature maps is represented as the convex combination of the two sets of feature maps:

\begin{equation} \label{eq:fm_fusion}
    \begin{aligned}
    \bm{{\rm {Conv}}} = \lambda \bm{{\rm {Conv}_1}} + (1-\lambda)\bm{{\rm {Conv}_2}},
    \end{aligned}
\end{equation}
where $\bm{{\rm {Conv}_1}}$ and $\bm{{\rm {Conv}_2}}$ represent feature maps generated by the two branches, respectively.

Accordingly, we define the overall loss function of DPLNet as follows:
\begin{equation} \label{eq:loss}
    \begin{aligned}
     \ell = \lambda \ell_{cls}(\bm{p}, \bm{y}^{(1)}) + (1 - \lambda) \ell_{cls}(\bm{p}, \bm{y}^{(2)}),
    \end{aligned}
\end{equation}
where $\lambda$ is the same as in Equation (\ref{eq:fm_fusion}), $\bm{p}$ is the predicted probability, and $\bm{y}^{(1)}$ and $\bm{y}^{(2)}$ are one-hot
encoding labels for images given to the first and the second input branch, respectively.
The cross entropy loss is used as classification loss $\ell_{cls}$ in this work.
If $0.5 <  \lambda < 1$, the DPLNet is more interested in learning  the corresponding image than the other image, and it receives more supervision for learning this image (see Equation (\ref{eq:loss})).
In this case, the other image can be seen as an auxiliary image for learning.
If $\lambda$ is equal to 0.5, the network has equal interest in  learning  two input images.
If $\lambda$ equals 0 or 1, the network  learns only from one image while disregarding the other image. In this case, the DPLNet degrades to a conventional deep network.

\begin{algorithm}[!t]
	\caption{Mini-batch stochastic gradient descent for training DPLNets.}
	\begin{algorithmic}[1]
		\FOR{$epoch=1$ to $nEpochs$}
            \FOR{$batch=1$ to $totalBatches$}
            \STATE inputs\_1, targets\_1 = get\_batch\_data()
            \STATE $\lambda$ = Uniform(0, 1)
            \STATE targets\_1, targets\_2 = random\_shuffle(inputs\_1, targets\_1)
            \STATE criterion = CrossEntropyLoss()
            \STATE optimizer.zero\_grad()
            \STATE outputs = net(inputs\_1, inputs\_2, $\lambda$)

            \STATE loss = $\lambda$criterion(outputs, targets\_1) \\ \quad $\,\;\;\;\;\;$ + $(1-\lambda)$criterion(outputs, targets\_2)
            \STATE loss.backward()

%
            \STATE optimizer.step()
            \ENDFOR
		\ENDFOR
	\end{algorithmic}
\label{alg:train}
\end{algorithm}

The DPLNet can be trained with the stochastic gradient descent (SGD) algorithm using a single data loader.
A pseudo  code snippet  is shown in Algorithm \ref{alg:train}.
In this algorithm, a batch data and their random shuffling are used for training.
This algorithm works equally well as using two data loaders.
It is more efficient than using two data loaders because it reduces I/O requirements.

\begin{figure}[!t]
\begin{center}
\includegraphics[width=0.618\linewidth]{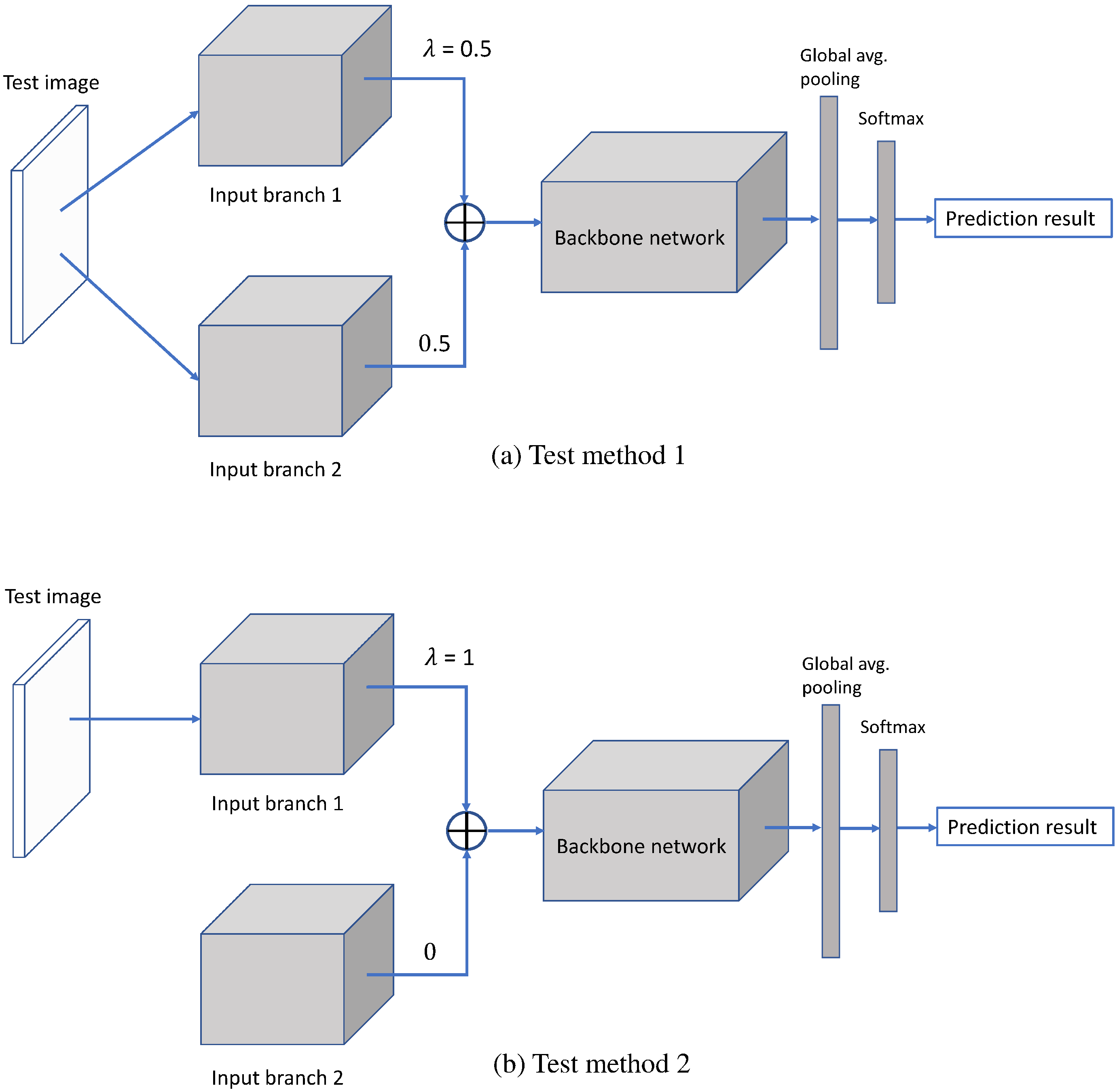}
\end{center}
   \caption{Two test approaches at test time: (a) Input a test image to both input branches and set $\lambda$ to 0.5; (b) Give a image to one input branch and set corresponding $\lambda$ to 1 while ignoring the other input branch.}

\label{fig:test_time}
\end{figure}

At test time, there are two approaches to test a image (see Figure \ref{fig:test_time}). The first approach is to give the test image as input to both  input branches, and set $\lambda$  to $0.5$.
The other approach is  to input the image to one input branch and set the corresponding $\lambda$ to 1 while ignoring the other input branch.
The final softmax layer produces a distribution over all categories.
Because we use convex combination of feature maps generated by two approaches as input to backbone network, the fused feature maps generated by the two approaches are the same; therefore, the two approaches  produce the same prediction result.

\subsection{Discussion}

DPLNets can be easily implemented by adapting  state-of-the-art deep networks, such as ResNets, DenseNets, and ResNeXts, while having the same number of parameters.
Compared with original networks, our DPLNets substantially reduces overfitting.
This is because of the following two reasons:
\begin{enumerate}
  \item  The dual input  architecture enables DPLNets to have a considerably large number of image pairs for training on the same dataset. Suppose the batch size for SGD is $B$, we have up to $B^2$ image pairs  from one batch data using Algorithm \ref{alg:train}; whereas the number is only $B$ for training original networks.
      Besides, DPLNets have the same number of parameters  as original  networks. Therefore, DPLNets can reduce overfitting  compared with original networks.
  \item We introduce to associate a random interest value  to each input branch for learning corresponding image during training. This method can be seen as a stochastic regularization technique, and can further improve the generalization performance.
\end{enumerate}

\section{Experiments} \label{sec:experiments}
We conducted experiments on a diverse of recognition tasks to show that the DPL framework is a general technique to improve the performance of deep networks.
We further evaluated DPLNets on small datasets to demonstrate their generalization performance.
\subsection{Image Classification} \label{sec:img_classification}
We conducted experiments on the CIFAR-10 and CIFAR-100 datasets (\cite{krizhevsky2009learning}).
The CIFAR-10 and CIFAR-100 datasets consist of  $32\times32$  color images drawn from 10 and 100 categories, respectively.
The training and testing sets contain 50,000 and 10,000 images, respectively.
We used PyTorch (\cite{paszke2017pytorch}) for implementation.
We implemented DPLNets based on four state-of-the-art deep networks, \emph{i.e.,} ResNets, DenseNets, pre-activation ResNets (PreAct ResNets) (\cite{he2016identity}), and ResNeXts to test the robustness of the proposed DPL framework.
We used the classification accuracies of the original  networks as baselines for comparison.
Each input branch of our DPLNets consists of one block, and the backbone networks consist of three blocks.
Following (\cite{he2016deep, he2016identity, huang2017densely}), we applied zero-padding of four pixels to  training images for training DPLNets based on ResNets, PreAct ResNets, and DenseNets. Following (\cite{xie2017aggregated}), we applied  zero-padding of eight pixels to training images for training DPLNets based on ResNeXts.
A $32\times32$ image was randomly cropped from the padded image or its horizontal flip as input data to train the networks. Each channel of input data were normalized to have zero mean and unit variance.
We did not use dropout, following the practice in (\cite{ioffe2015batch}).
All models were trained from scratch using SGD for 300 epochs with a mini-batch of  128 examples.
The learning rate started from 0.1 and was divided by 10 at epoch 150 and 225.
The values of weight decay and momentum were set to 0.0005 and 0.9, respectively.
At test time, we only evaluated the original $32\times32$ image.

\subsubsection{DPLNets based on  ReseNets}
We implemented DPLNets based on five ResNet architectures, \emph{i.e.,} ResNet-18, ResNet-34, ResNet-50, ResNet-101, and ResNet-152.
We trained the models on a single GPU.
The experimental results are shown in Table \ref{tab:dpl_resnet}.
From this table, we observe  that as with ResNets,  the performance of our DPLNets improves as the number of layers increases.
The DPLNet based on PreAct ResNet-152 achieves the best performance on the two datasets,  whereby it achieves 3.75\% and 18.30\% error rate, respectively.
Using DPL  improves the recognition accuracy for the five ResNet architectures, with at least 0.31\% and 1.29\% performance improvements on the two datasets, respectively.
On average, DPLNets achieve 0.32\% and 1.84\% performance gains on the two datasets, respectively.
The performance improvements yielded by using DPL are higher on CIFAR-10 than on CIFAR-100.

\begin{table}[!ht]
\centering
\begin{tabular}{|c|c|c|}  
\hline
Model &CIFAR-10  &CIFAR-100\\
%
%
\hline
\hline
ResNet-18 & 4.96& 22.78 \\
ResNet-18 + DPL & \textbf{4.69}& \textbf{20.75} \\
\hline
\hline
ResNet-34  &4.86& 21.64 \\
ResNet-34 + DPL & \textbf{4.56}& \textbf{20.35} \\
\hline
\hline
ResNet-50 & 4.62& 21.89 \\
ResNet-50 + DPL  & \textbf{4.29}& \textbf{19.32} \\
\hline
\hline
ResNet-101 & 4.44& 20.81 \\
ResNet-101 + DPL  & \textbf{4.04}& \textbf{19.07} \\

\hline
\hline
ResNet-152 & 4.32& 19.95 \\
{ResNet-152 + DPL} & \textbf{4.01}& \textbf{18.39} \\
\hline
\end{tabular}
\captionof{table}{Test errors (\%) on CIFAR-10 and CIFAR-100.} 
\label{tab:dpl_resnet}
\end{table}

\subsubsection{DPLNets based on PreAct ReseNets}
For experiments of DPLNets based on PreAct ResNets, we explored five PreAct ResNet architectures, \emph{i.e.,} PreAct ResNet-18, PreAct ResNet-34, PreAct ResNet-50, and PreAct ResNet-101..
We trained the models on a single GPU.
The experimental results are shown in Table \ref{table:results_preact}.
From this table, we find  that as with PreAct ResNets,  the performance of  DPLNets improves as the number of layers increases.
Our DPLNet based on PreAct ResNet-152 achieves the best performance on the two datasets,  whereby it achieves 3.75\% and 18.30\% error rates, respectively.
As with DPLNets based on ResNets, using DPL  improves the classification accuracy for the five PreAct ResNet architectures.
Using DPL yields at least 0.42\% and 1.54\% performance improvements on the two datasets, respectively.
On average, our DPLNets achieves 0.52\% and  1.73\% performance improvements on the two datasets, respectively.
The DPLNets exhibate better performance based on PreAct ResNet  than  based on ResNet.

\begin{table}[!ht]
\centering  
\begin{tabular}{|c|c|c|}  
\hline
Model &CIFAR-10  &CIFAR-100\\
\hline
\hline
PreAct ResNet-18 & 4.90& 22.48 \\

PreAct ResNet-18 + DPL& \textbf{4.16}& \textbf{20.15} \\
\hline
\hline
PreAct ResNet-34  & 4.69& 21.08 \\

PreAct ResNet-34 + DPL& \textbf{4.12}& \textbf{19.54} \\
\hline
\hline
PreAct ResNet-50 & 4.52& 20.60 \\
PreAct ResNet-50 + DPL& \textbf{4.07}& \textbf{19.01} \\
\hline
\hline
PreAct ResNet-101 & 4.38& 20.51 \\

PreAct ResNet-101 + DPL& \textbf{3.96}& \textbf{18.89} \\
\hline
PreAct ResNet152 & 4.18& 19.87 \\

{PreAct ResNet-152 + DPL}& \textbf{3.75}& \textbf{18.30} \\
\hline
\end{tabular}
\caption{Test errors (\%) on CIFAR-10 and CIFAR-100. } 
\label{table:results_preact}
\end{table}

\subsubsection{DPLNets based on DenseNets}
For experiments of using DLP based on DenseNets, we investigated two DenseNet architectures, \emph{i.e.,} DenseNet-121 and DenseNet-169.
The growth rate $k$ was set to 32 in our experiments.
The models were trained on a single GPU.
The performance comparison of the DPLNets and original DenseNets  is shown in Table \ref{tab:dpl_densenet}.
From this table, we see that our DPLNets  achieve better performance than original networks again.
The DPLNet based on DenseNet-169 achieves the highest recognition accuracy on the two datasets, wherein it achieves 4.31\% and 18.36\% error rates, respectively.
Our DPLNets yield at least 0.15\% and 1.85\% performance improvement on the two datasets, respectively.

\begin{table}[!ht]
\centering
\begin{tabular}{|c|c|c|}  
\hline
Model &CIFAR-10  &CIFAR-100\\
\hline
\hline
DenseNet-121 ($k=32$)& 4.55& 22.0 \\

DenseNet121 + DPL & \textbf{4.43}& \textbf{19.11} \\
\hline
\hline
DenseNet-169 ($k=32$)  & 4.46& 20.21 \\

{DenseNet-169 + DPL}& \textbf{4.31}& \textbf{18.36}\\
\hline

\hline
\end{tabular}
\caption{Test errors (\%) on CIFAR-10 and CIFAR-100. $k$ indicates network's growth rate.} 
\label{tab:dpl_densenet}
\end{table}

\subsubsection{DPLNets based on ResNeXts}
We implemented DPLNets based on two ResNeXt architectures, \emph{i.e.,} ResNeXt-29 (8$\times$64d) and ResNeXt-29 (16$\times$64d).
The number of channels in each group is 64.
The models with cardinality equal to 8 were trained on two GPUs, and the models with cardinality equal to 16 were trained on four GPUs.
The comparison of results of our DPLNets and original ResNeXts  is show in Table \ref{table:dpl_ResNeXt}. From this table, we find that our DPLNets achieve better performance than original ResNeXts.
The DPLNet based on ResNeXt-29 (16$\times$64d) achieves the best performance on the two datasets, wherein it achieves 3.44\% and 16.90\% error rates, respectively. The performance improvements yielded by using DPL on CIFAR-10  is not significant compared with on CIFAR-100.

\begin{table}[!ht]
\centering  
\begin{tabular}{|c|c|c|}  
\hline
Model &CIFAR-10  &CIFAR-100\\
\hline
ResNeXt-29 $8\times64$d \cite{xie2017aggregated}& 3.65& 17.77 \\

ResNeXt-29 $8\times64$d  + DPL& \textbf{3.48}& \textbf{17.15} \\
\hline
\hline
ResNeXt-29 $16\times64$d \cite{xie2017aggregated}& 3.58& 17.31 \\

{ResNeXt-29 $16\times64$d + DPL} & \textbf{3.44}& \textbf{16.90} \\ 
\hline
\end{tabular}
\caption{Test errors (\%) on CIFAR-10 and CIFAR-100.}
\label{table:dpl_ResNeXt}
\end{table}

\subsubsection{Discussion}
We have seen that using the proposed DPL framework improves classification accuracies for the four types of deep network architectures, \emph{i.e.,} ResNets, DenseNets, PreAct ResNets, and ResNeXts.
This shows that the performance of  DPLNets is stable.
It is worth noting that the DPLNets have the same number of parameters as original networks.
We observe that the performance improvements achieved by DPLNets are more  significant on CIFAR-100 than on CIFAR-10.
In CIFAR-10 and CIFAR-100, each category has 5000 and 500 samples, respectively.
The results indicate that  DPLNets are very helpful for small training sets.

\begin{figure*}[!t]
\begin{center}

\includegraphics[width=.818\linewidth]{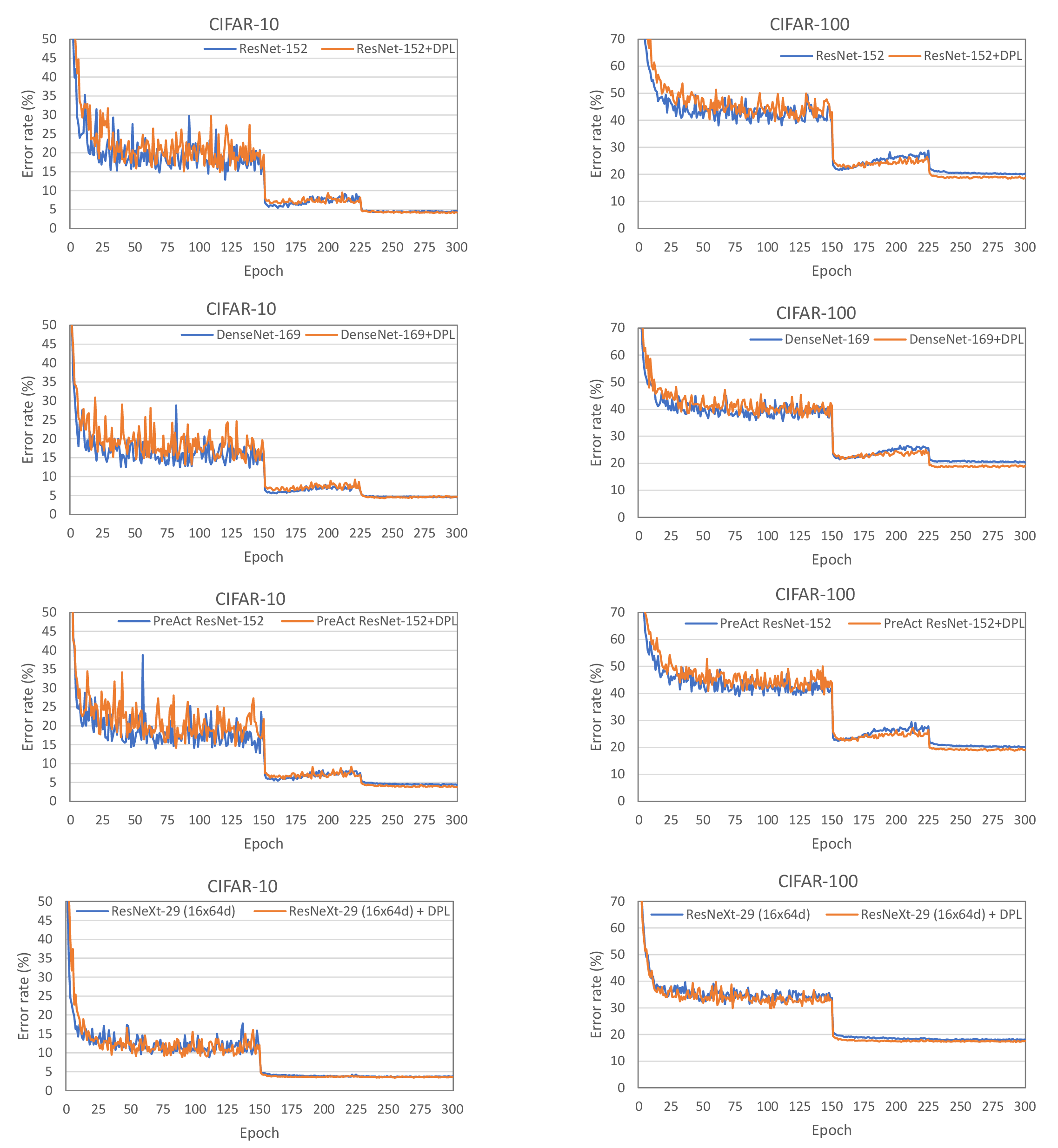}
\end{center}
   \caption{Test error evolutions for DPLNets and original networks on CIFAR-10 and CIFAR-100. The error rates of DPLNets (except the DPLNet based on ResNeXt on CIFAR-100) converge slowly from epoch 1 to epoch 149 compared with original networks. }
\label{Fig:error_rate_overall}

\end{figure*}

Figure \ref{Fig:error_rate_overall} shows  test error evolutions for  DPLNets and original networks.
We can see that test  error evolutions of DPLNets based on ResNet, PreAct ResNet, and DenseNet converge slowly compared with original network at the beginning (from epoch 1 to epoch 149).
This is because DPLNets need more epochs to converge to learn discriminative class-specific features.
The DPLNets exhibit comparable performance as original networks at the middle stage (from epoch 150 to epoch 224).
Our DPLNets finally achieve lower error rates than original network at the last stage (from epoch 225 to epoch 300).

\subsection{Image Emotion Classification}

For image emotion recognition, experiments were carried out on the FI-8 dataset (\cite{you2016building}). This dataset was collected from Flickr and Instagram.
There are totally 23,308 images labelled with eight emotion categories. The FI-8 dataset is randomly split  into 80\%
training, 5\% validation, and 15\% testing sets. 
In our experiments, all training images were resized with the size of the shorter side equal to 256 while maintaining the original aspect ratio.
A 224$\times$224 image was randomly cropped from original image or its horizontal flip as input data to networks.
Each channel of input data was normalized to have zero mean and unit variance.
At  test time, the network made a prediction by  cropping 10 regions of the size of $224\times224$ (four corners and one center, and their horizontal flip) from a test image, and averaging the predictions made by the network's softmax layer on the ten patches.

We implemented DPLNets based on ResNets, which are pre-trained on ImageNet.
Each input branch of the DPLNets consists of one block, and the backbone networks consist of three blocks.
We trained the DPLNets using SGD for $90$ epochs with a mini-batch of size 128.
The values of weight decay and momentum were set to 0.0001 and 0.9, respectively. The learning rate started from 0.1 and was divided by 10 after 30 and 60 epochs.
The experimental results are shown in Table \ref{tab:dpl_fi}.
From this table, we observe that as with ResNets,
the performance of ResNets and DPLNets improves as the
number of layers increases.
The DPLNets achieve better performance than original ResNets. The DPLNet based on ResNet-152 achieves the best classification accuracy of 68.18\% on this dataset. Using DPL yields an average of  1.55\% performance improvement.

\begin{table}[!ht]
\centering  
\begin{tabular}{|c|c|}  
\hline
Model & Recognition accuracy (\%) \\
\hline
\hline
ResNet-18 & 64.48 \\

ResNet-18 + DPL &  \textbf{65.58}\\
\hline
\hline
ResNet-34 & 65.08 \\

ResNet-34 + DPL&  \textbf{66.76}\\
\hline
\hline
ResNet-50 &  65.99\\

ResNet-50 + DPL&  \textbf{67.90 }\\
\hline
\hline
ResNet-101 & 66.56 \\
ResNet-101 + DPL& \textbf{68.18 }\\
\hline
\hline
ResNet-152 & 67.07 \\
ResNet-152 + DPL& \textbf{68.51} \\

\hline
\end{tabular}
\caption{Recognition accuracies (\%)  on FI-8.}
\label{tab:dpl_fi}
\end{table}

\subsection{MNIST classification} \label{sec:mnist}
\begin{figure}[!t]
\begin{center}
\includegraphics[width=.668\linewidth]{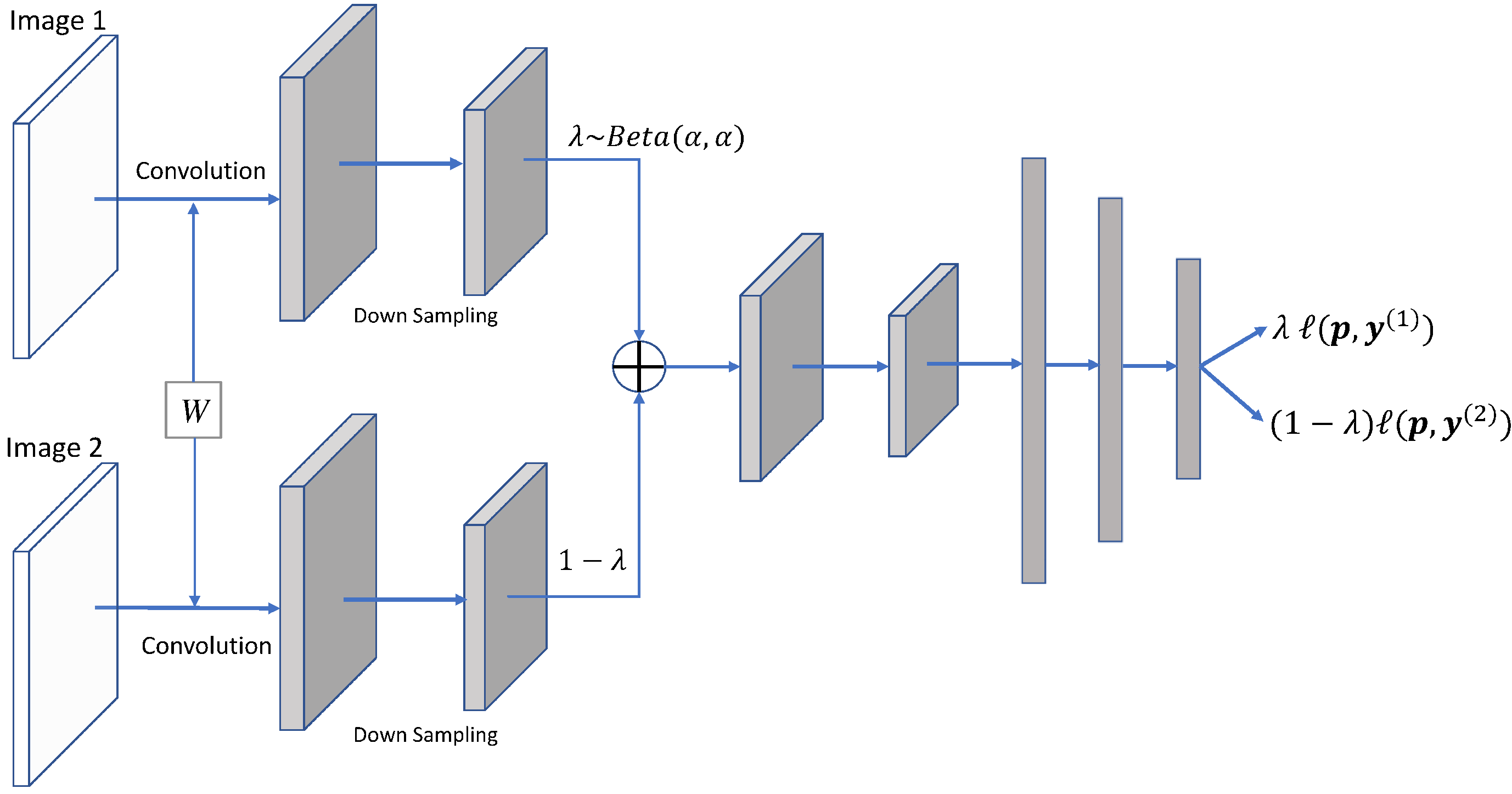}
\end{center}
\caption{DPLNet based on LeNet-5.}
\label{fig:dpl_lenet5}
\end{figure}
The MNIST digit dataset consists of 60,000 training and 10,000 testing images of ten handwritten digits (0 to 9), each with $28\times28$ pixels.
We implemented DPLNet based on  LeNet-5 (\cite{lecun1998gradient}).
The LeNet-5 consists of two convolutional layers, which are followed by subsampling layers, and three fully connected layers with a final softmax.
In our implementation using DPL, each input branch contains a convolutional and a subsampling layer, and the backbone network consists of a convolutional layer, a subsampling layer and three fully connected layers (see Figure \ref{fig:dpl_lenet5}).
The model was trained from scratch using Adam for 500 epochs with a mini-batch of 128 examples. The learning rated was set to 0.001.

\begin{table}[!ht]
\centering
\begin{tabular}{|c|c|}  
\hline
Model  &Test set\\
\hline
\hline
LeNet-5 [no distortions] \cite{lecun1998gradient} & 0.95 \\

LeNet-5 [huge distortions] \cite{lecun1998gradient} & 0.85 \\

LeNet-5 [distortions] \cite{lecun1998gradient} & 0.8 \\

LeNet-5 + DPL  & \textbf{0.55} \\
\hline
\end{tabular}
\caption{Error rates (\%) on MNIST.}
\label{tab:dpl_mnist}
\end{table}

The experimental results are shown in Table \ref{tab:dpl_mnist}.
From this table, we find DPL achieves 0.28\% higher performance than original LeNet-5.
Our approach also achieve better performance than LeNet-5 with distortions of the input.

\subsection{Google commands dataset}
We further conducted experiments  on speech data. 
We used the Google commands dataset (Warden). This dataset consists of 65,000 utterances which were recorded by thousands of different people.
There are totally 30 categories.
Each utterance is about one-second long and belongs to one out of 30 short words, such as \emph{yes, no, down,
left}. Following (\cite{zhang2017mixup}), we down-sampled from the original waveforms with the sampling rate equal to 16 kHz, and extracted normalized spectrograms.
We applied zero-padding to the spectrograms such that their sizes equal to $160 \times 101$.
We implemented DPLNet based on LeNet-5 (\cite{lecun1998gradient}).
Each input branch consists of a convolutional and a subsampling layer, and feature maps generated by two input branches are then fused to backbone network (see Figure \ref{fig:dpl_lenet5}).
The first  fully connected layer contains 16280  neurons, and the second fully connected layer contains 1000 neurons.
The models were trained using SGD with a mini-batch of 100 examples.
The learning rated started at 0.001 and was divided by 10 after 50 epochs. The experimental results are shown in Table \ref{table:googlecmd}.
From this table, we see that our DPLNet yields 1.2\% and 1.5\% performance improvement on the validation set and the testing set, respectively.

\begin{table}[!ht]
\centering
\begin{tabular}{|c|c|c|}  
\hline
Model &Validation set  &Test set\\
\hline
\hline
LeNet-5  & 9.8& 10.3 \\

LeNet-5 + DPL & \textbf{8.6}& \textbf{8.8} \\
\hline
\end{tabular}
\vspace{5pt}
\caption{Error rate (\%) on the Google commands dataset.}
\label{table:googlecmd}

\end{table}

\subsection{Experiments on Small Datasets}
We conducted experiments on small datasets to further evaluate the generalization performance of our DPLNets.
We used subsets of CIFAR-10, CIFAR-100, and MNIST. For experiments on subsets of CIFAR-10 and CIFAR-100, we randomly selected 100, 200, 300, 400, and 500 training samples from each category. The DPLNets were implemented based on ResNet-18 and PreAct ResNet-18.
The parameter setting and the training procedures were the same as in section \ref{sec:img_classification}.
For experiments on subsets of MNIST, we randomly selected 10, 20, 30, 50, and 100 training samples from each category. We used the same DPLNet structure (see Figure \ref{fig:dpl_lenet5}), parameter setting, and training procedures  as in section \ref{sec:mnist}. The experimental results are shown in Table \ref{tab:small_cifar10}, Table \ref{tab:small_cifar100}, and Table \ref{table:mnist_small}, respectively.

\begin{table*}[!ht]
\centering  
\begin{tabular}{|c|c|c|c|c|c|}  
\hline
Num. of Samples  per Category & 100 & 200 & 300 &400 & 500 \\
\hline
\hline
ResNet-18 &45.76&34.70 &27.57&21.07&18.46 \\
\textbf{ResNet-18 + DPL }     &\textbf{40.78} &\textbf{26.96} &\textbf{20.43}&\textbf{l7.56}&\textbf{15.79}\\
\hline
\hline
PreAct ResNet-18 &45.32&33.61&25.62&20.71& 18.38\\
\textbf{PreAct ResNet-18 + DPL}         &\textbf{36.71}&\textbf{24.58}&\textbf{19.57}&\textbf{17.04}&\textbf{15.16}\\
\hline
\end{tabular}
\caption{Error rates (\%)  on subsets of CIFAR-10.}
\label{tab:small_cifar10}
\end{table*}

\begin{table*}[!ht]
\centering  
\begin{tabular}{|c|c|c|c|c|c|}  
\hline
Num. of Samples  per Category & 100 & 200 & 300 &400 & 500 (Full dataset) \\
\hline
\hline
ResNet-18 &55.35&67.42 &72.93&75.78&22.78 \\
\textbf{ResNet-18 + DPL }     &\textbf{59.37} &\textbf{70.86} &\textbf{75.37}&\textbf{78.15}&\textbf{20.57}\\
\hline
\hline
PreAct ResNet-18 &56.05&66.97&71.84&74.95&22.48 \\
\textbf{PreAct ResNet-18 + DPL}         &\textbf{60.39}&\textbf{71.07}&\textbf{75.51}&\textbf{77.52}&\textbf{20.51}\\
\hline
\end{tabular}
\caption{Error rates (\%)  on subsets of CIFAR-100.}
\label{tab:small_cifar100}
\end{table*}

\begin{table*}[!ht]
\centering  

\begin{tabular}{|c|c|c|c|c|c|}  
\hline
Num. of Samples  per Category & 10 & 20 & 30 &50 & 100 \\
\hline
\hline
LeNet-5 &25.49&18.10&14.05&8.80&5.76 \\
\textbf{LeNet-5 + DPL}          &\textbf{20.14}&\textbf{14.73} &\textbf{9.67}&\textbf{7.33}&\textbf{4.29}\\
\hline
\end{tabular}
\caption{Error rates (\%)  on subsets of MNIST.}
\label{table:mnist_small}
\end{table*}

From Table \ref{tab:small_cifar10}, we find that the DPLNet based on ResNet-18 and the DPLNet based on PreAct ResNet-18 yield the heighest performance improvements of 7.74\% and 9.03\%, respectively, with each category  has 200 training samples on CIFAR-10 compared with original networks.
As the number of training samples in each category increases from 300 to 500, the performance improvement decreases.
Overall, they achieve at least 2.67\% and 3.22\% performance improvements, respectively.
The DPLNets achieve an average of 2.89\% and 3.33\% performance improvement on subsets of CIFAR-100, respectively.
The performance improvements are higher on CIFAR-10 than on CIFAR-100.
This is because CIFAR-100 has more categories, and this makes networks are difficult to train on this dataset.
Our DPLNet yields  an average of 3.21\% performance improvement (see Table \ref{table:mnist_small}).

A general observation from the three tables is that the performance improvement yielded by DPLNets is high on subsets with each category has a small number of samples.
The experimental results show that the proposed dual patter learning framework is very helpful for small subsets. The dual pattern learning architecture might be promising for other tasks in which training samples are extremely to collect.

\section{Conclusion} \label{sec:conclusion}

In this paper, we have presented the dual pattern learning network architecture which is characterized by two input branches and two loss functions.
We showed that DPLNets which are trained by minimizing the empirical dual prediction risk of a training dataset are effective for single image classification.
This architecture can learn class-specific features by analyzing and comparing dual inputs compared with conventional networks.
In addition, the dual input structure enables the network to have a large number of image pairs to train the network, which can help address the overfitting issue due to lack of training data.
Furthermore, we introduced a stochastic regularization method which can further improve the generalization performance of DPLNets.
We  evaluated DPLNets on on a diverse of classification tasks including image classification, image emotion recognition, handwritten digit recognition  and speech recognition. The experimental results showed that DPLNets could lead to performance improvement over state-of-the-art networks.

The proposed empirical dual prediction risk minimization method is not restricted to image classification. It could be applied to other recognition tasks, such as sequence classification and object detection/segmentation. This would be investigated in our future work.

\bibliography{myref}

\end{document}